\documentclass{article}

\PassOptionsToPackage{numbers, compress}{natbib}

\usepackage[final]{ewrl_2026}

\usepackage[utf8]{inputenc}
\usepackage[T1]{fontenc}
\usepackage{hyperref}
\usepackage{url}
\usepackage{booktabs}
\usepackage{amsmath}
\usepackage{amssymb}
\usepackage{amsfonts}
\usepackage{amsthm}
\usepackage{microtype}
\usepackage{xcolor}
\usepackage{graphicx}

\newcommand{\D}{\mathcal{D}}

\newcommand{\E}{\mathbb{E}}

\newcommand{\samp}{\textsc{Sample}}
\newcommand{\avg}{\textsc{Avg}}
\newcommand{\model}{\textsc{Model}}

\newcommand{\mpTail}{\textsc{mp-tail}}
\newcommand{\se}{\,{\scriptstyle\pm}\,}

\title{Correcting Within-Group Self-Selection Bias in Prioritized Replay}

\author{
  Oscar Mir\'o L\'opez-Feliu \\
  University of Amsterdam \\
  {\small\texttt{oscar.miro.lopez.feliu@student.uva.nl}}
  \And
  Herke van Hoof \\
  University of Amsterdam \\
  {\small\texttt{h.c.vanhoof@uva.nl}}
}

\begin{document}
\maketitle

\begin{abstract}
Prioritized experience replay (PER) improves sample efficiency by replaying high-priority transitions, usually according to absolute temporal-difference error. In stochastic environments, PER can distort the distribution of realized outcomes replayed from transitions with the same state-action pair. We call this within-group self-selection. We quantify the resulting changes in within-group outcome frequencies and mean Bellman targets. We decompose PER into between-group allocation and conditional sibling selection, and derive fixed-buffer corrections that preserve current group-level priority mass: \samp\ selects a group through PER and trains on a uniformly sampled sibling; \avg\ averages sibling Bellman targets; and \model\ samples from an empirical full-outcome model. In exact state-action environments with rare high-magnitude outcomes, sibling-aware replay improves learning efficiency over PER, although matched parameter sweeps show that tuning can narrow some gaps. In MinAtar, approximate VQ-VAE groups with \samp\ mitigate degradation under mean-preserving reward tails in four of five games. Sibling-aware replay thus retains the focus on high-priority state-action regions while recovering their empirical outcome frequencies.

\end{abstract}

\section{Introduction}
Experience replay stores past transitions and reuses them for updates, reducing temporal correlation and improving data reuse in off-policy reinforcement learning \citep{Lin1992}. In deep value learning, DQN combined replay with target networks to stabilize training from high-dimensional observations \citep{Mnih2015}, while prioritized experience replay (PER) changed the replay distribution by sampling transitions according to priority, typically a power of absolute temporal-difference (TD) error \citep{Schaul2016}. Priority-driven replay has since become part of high-performing value-based agents such as Rainbow, Ape-X, and R2D2 \citep{Hessel2018,Horgan2018,Kapturowski2019}. More recent analyses show that replay capacity, replay ratio, and buffer composition can substantially change learning outcomes \citep{ZhangSutton2017,Fedus2020}. For PER, transition-level prioritization affects both which state-action regions are revisited and which stochastic outcomes dominate within those regions.

PER commonly applies importance-sampling (IS) weights to correct its non-uniform update distribution toward uniform replay \citep{Schaul2016}. In practice, this correction is often partial because the IS exponent is annealed toward one. Normalizing weights within the sampled minibatch can also change the expected update. Stale priorities and moving targets further affect learning dynamics \citep{Pan2022}. Prior work studies the effective loss, coverage, and gradient-estimation consequences of non-uniform replay \citep{Fujimoto2020,Pan2022,Lahire2022,Li2021}; other methods change priorities to reduce replay driven by irreducible or aleatoric noise \citep{Sujit2023,CarrascoDavis2025}. Together, these works show why transition priority can be useful yet imperfect as a proxy for learning value.

This paper isolates a conditional form of replay distortion. Consider a replay buffer containing several transitions with the same state-action pair $(s,a)$ but different realized stochastic outcomes. We call these transitions \emph{siblings}. Throughout, a \emph{tail} means a low-probability, high-magnitude outcome relative to the main mass of a conditional reward or target distribution. PER performs two operations at once. First, it allocates more replay mass to some state-action groups than others; we call this \emph{between-group allocation}. Second, within a selected group, it changes which realized outcomes are replayed more often; we call this \emph{within-group self-selection}. Prior work mainly changes the transition priority score, and therefore changes the overall transition distribution or the amount of replay mass assigned to different regions. We instead keep the group-level priority mass and target the second effect: the conditional outcome frequencies used once a group has already been selected.

The first effect is often the desired behavior of PER, since some state-action regions may deserve more updates. The second effect can be harmful. If rare high-magnitude outcomes have persistently large TD errors for aleatoric reasons, then PER can replay the right region at the wrong conditional outcome frequencies. In that case, the problem is not that some transitions receive too much replay probability globally, but that the empirical outcome distribution inside a fixed state-action group is tilted toward high-priority realizations.

We propose sibling-aware replay methods that separate PER's group-level and within-group effects. At a fixed buffer and current priorities, the group-level priority mass is preserved, so high-priority state-action regions can still receive more replay effort. Within a selected group, the update is instead constructed from the empirical sibling distribution. \samp\ trains on a uniformly sampled sibling transition; \avg\ averages sibling Bellman targets; and \model\ samples from the empirical full-outcome distribution for the exact state-action pair. Our contributions are: 1) We formalize a finite-buffer decomposition of PER into priority mass over state-action groups and priority-weighted sibling selection within each group. 2) We derive the induced sampling distribution and group-aware importance weights for \samp, and we define \avg\ and \model\ as empirical estimators of the within-group Bellman target. 3) We evaluate the resulting methods in exact stochastic environments, where sibling groups are exact state-action groups, and in a function-approximation setting where frozen VQ-VAE codes provide approximate latent groups.

\section{Related work}
\paragraph{Experience replay and prioritized value learning.}
Replay reuses past experience \citep{Lin1992} and became a core component of deep value learning through DQN \citep{Mnih2015}. PER replaces uniform transition sampling with priority-based sampling and importance weights \citep{Schaul2016}; it also appears in approaches such as Rainbow, Ape-X, and R2D2 \citep{Hessel2018,Horgan2018,Kapturowski2019}. Replay capacity, replay ratio, and buffer composition can substantially affect learning \citep{ZhangSutton2017,Fedus2020}. These findings motivate studying replay as an algorithmic sampling distribution; we examine its conditional structure within state-action groups.

\paragraph{Analyses of prioritized replay.}
\citet{Fujimoto2020} relate non-uniform sampling to loss modification and analyze PER's outlier bias in conditional value targets. \citet{Pan2022} connect error-based prioritization to an effective loss and identify stale priorities and insufficient coverage as limitations. LaBER formulates replay sampling as a gradient-estimation problem \citep{Lahire2022}, while \citet{Li2021} analyze when TD error is a good proxy for the value of replaying a transition. ROER derives replay reweighting from regularized occupancy optimization toward an optimal on-policy distribution \citep{Li2024ROER}; DisCor reweights using estimated target accuracy to improve corrective feedback under bootstrapping \citep{Kumar2020DisCor}. These objectives differ from IS correction toward uniform replay. Our decomposition instead separates the probability of selecting a state-action group from the outcome distribution within that group.

\paragraph{Noisy, irreducible, and aleatoric transitions.}
High TD error can reflect noise rather than a useful learning signal. ReLo uses online--target-network loss differences as a proxy for reducible loss \citep{Sujit2023}; UPER prioritizes information gain estimated from epistemic and aleatoric uncertainty \citep{CarrascoDavis2025}. These methods are close conceptual neighbors because they address when high-error transitions are useful replay targets. They change transition priority scores, which can affect both group allocation and conditional outcome selection. Our correction instead retains the current priority mass over groups and recovers empirical sibling frequencies within each selected group, without classifying error sources. These approaches are complementary: noise-aware priorities could improve group allocation, while sibling-aware sampling controls the outcome frequencies used within groups.

\section{Finite-buffer decomposition of PER}
Let a replay buffer contain $N$ transitions
\begin{equation}
\D = \{z_i\}_{i=1}^N, \qquad z_i = (s_i,a_i,r_i,s'_i,d_i^{\mathrm{term}},d_i^{\mathrm{trunc}}),
\end{equation}
where $d_i^{\mathrm{term}}$ and $d_i^{\mathrm{trunc}}$ denote termination and truncation indicators. In Bellman targets, $d_i$ denotes the effective no-bootstrap indicator, equal to their maximum in our experiments. Let $p_i>0$ be the stored raw priority assigned by the replay rule; in our experiments, priority updates use $p_i=|\delta_i|+\varepsilon$, where $\delta_i$ is the pre-update TD error and $\varepsilon>0$ prevents zero sampling probability. Let $\alpha\geq 0$ be the priority exponent and define the sampling mass $u_i=p_i^\alpha$. PER samples transition $i$ with
\begin{equation}
P_{\mathrm{PER}}(i)=\frac{u_i}{S}, \qquad S=\sum_{j=1}^{N} u_j,
\label{eq:perlaw}
\end{equation}
where $S$ is the total priority mass in the buffer. For fixed priorities and a fixed per-transition loss $\ell_i(\theta)$, the idealized expected objective under unnormalized IS weights $w_i=(N P_{\mathrm{PER}}(i))^{-\beta}$ is
\begin{equation}
\mathcal{J}_{\mathrm{PER},\beta}(\theta)
=
\sum_{i=1}^{N} P_{\mathrm{PER}}(i) w_i\, \ell_i(\theta)
\propto
\sum_{i=1}^{N} u_i^{1-\beta}\ell_i(\theta),
\label{eq:weightedobj}
\end{equation}
where $\beta\in[0,1]$ controls the strength of correction and the omitted factor is independent of $\theta$ when priorities are held fixed \citep{Fujimoto2020,Pan2022}. With unnormalized weights, $\beta=1$ recovers the uniform-buffer expected gradient, while $\beta<1$ leaves residual priority weighting \citep{Schaul2016}. The analysis below uses unnormalized weights; our implementation uses minibatch-maximum normalization, whose effect is discussed in Appendix~\ref{app:derivations}.

Now define a fixed group key $\kappa(i)$. In the exact tabular setting, $\kappa(i)=(s_i,a_i)$. A group $g$ contains the sibling indices
\begin{equation}
I(g)=\{i:\kappa(i)=g\}, \qquad n_g=|I(g)|, \qquad s_g=\sum_{j\in I(g)} u_j .
\end{equation}
Here, $n_g$ is the number of observed siblings in group $g$, and $s_g$ is their total priority mass. The PER sampling distribution decomposes as
\begin{equation}
P_{\mathrm{PER}}(\kappa=g)=\frac{s_g}{S},
\qquad
P_{\mathrm{PER}}(i\mid \kappa=g)=\frac{u_i}{s_g}, \quad i\in I(g).
\label{eq:decomp}
\end{equation}
Equation~\eqref{eq:decomp} separates two phenomena. The first term is PER's between-group allocation: some state-action groups receive more updates. The second is within-group sibling selection: once a group is revisited, high-priority realized outcomes are over-sampled relative to their empirical frequency. The empirical conditional sibling distribution is instead $P_{\D}(i\mid \kappa=g)=1/n_g$.

Normalizing the expected weighted coefficients in Equation~\eqref{eq:weightedobj} within $g$ gives
\begin{equation}
\pi_{\beta}(i\mid g)=\frac{u_i^{1-\beta}}{\sum_{j\in I(g)}u_j^{1-\beta}}.
\label{eq:effective-exponent}
\end{equation}
This is an effective within-group weighting, not the actual conditional sampler, which remains $u_i/s_g$. Reducing $\alpha$ moves transition sampling toward uniform replay, affecting both components; increasing $\beta$ changes update weights, leaving the sampler unchanged at fixed priorities. Both can indirectly change future priorities. At $\alpha=0$, PER and \samp\ both sample uniformly from the buffer. Appendix~\ref{app:derivations} details the assumptions and normalization distinction.

The same distinction appears in the update target. For any fixed scalar target $T_i$ associated with transition $i$, the empirical sibling target for group $g$ is the uniform average
\begin{equation}
\bar T_g = \frac{1}{n_g}\sum_{i\in I(g)} T_i,
\end{equation}
whereas PER without IS correction ($\beta=0$) uses the priority-weighted conditional mean
\begin{equation}
\tilde T^{\mathrm{PER}}_g
=
\sum_{i\in I(g)} \frac{u_i}{s_g} T_i .
\end{equation}
We call the gap between $u_i/s_g$ and $1/n_g$ \emph{within-group self-selection}. It is a finite-buffer sampling property, not by itself a claim about the true environment distribution. Uniform sibling sampling exactly recovers the empirical conditional distribution over observed siblings; it approximates the environment conditional distribution only insofar as the buffer is representative. The pathology appears when priority is driven by rare stochastic outcomes rather than reducible learning error: PER then biases the update toward high-priority realizations inside the group, rather than the empirical outcome frequencies.

\paragraph{Scope.}
All exact statements below assume a fixed buffer, fixed priorities, and fixed target parameters. They characterize the stochastic update distribution induced by replay. The decomposition applies to any fixed key map $\kappa$; our exact claims use $\kappa(i)=(s_i,a_i)$, while the function-approximation experiments later use latent keys of the form $\kappa_\phi(i)=(\phi(s_i),a_i)$ for a frozen abstraction $\phi$. Latent keys correct only within the chosen abstraction and can introduce abstraction error.

\section{Sibling-aware replay}
The methods below preserve PER's group-selection role but change the outcome used inside the selected group. We treat the transition $A$ selected by PER as an anchor that determines the sibling group $g=\kappa(A)$. A within-group rule then constructs the transition or target used in the update. Appendix~\ref{app:derivations} gives the corresponding finite-buffer derivations.

\subsection{SAMPLE: uniform sibling replacement}
\samp\ treats the PER anchor only as a group selector. After drawing $A\sim P_{\mathrm{PER}}$, it samples the trained index $J$ uniformly from the anchor's sibling group $I(\kappa(A))$ and trains on the full transition $z_J$. Thus, for any trained index $j$, writing $g=\kappa(j)$,
\begin{equation}
q_{\mathrm{SAMPLE}}(j)
=
\Pr(J=j)
=
\sum_{i\in I(g)} \frac{u_i}{S}\frac{1}{n_g}
=
\frac{s_g}{S n_g}.
\label{eq:samplelaw}
\end{equation}
The finite-buffer IS ratio from this sampler back to uniform replay is
\begin{equation}
\rho_j
=
\frac{1/N}{q_{\mathrm{SAMPLE}}(j)}
=
\frac{n_gS}{Ns_g},
\qquad
w_j^{\mathrm{grp}}=\rho_j^\beta .
\label{eq:groupweights}
\end{equation}
At $\beta=1$, fixed priorities, and fixed per-transition losses, \samp\ with \emph{unnormalized} group-aware IS matches the uniform-buffer expected estimator. For $\beta<1$, residual weighting remains across groups, but siblings are uniform within each group for every $\beta$. Group weights, including their minibatch maximum, depend only on selected groups and therefore do not reintroduce conditional sibling weighting. These are fixed-buffer identities: different trained transitions and priority refreshes can make group masses diverge between methods during learning.

\subsection{AVG and MODEL}
In our value-based experiments, let
\begin{equation}
Y_j =
r_j + \gamma(1-d_j)
Q_{\bar\theta}\!\left(s'_j,\arg\max_a Q_\theta(s'_j,a)\right)
\end{equation}
be the Double-DQN target for sibling $j$ \citep{VanHasselt2016}. \avg\ is a target-level correction: it replaces the sampled target by an average of sibling Bellman targets,
\begin{equation}
\widehat{Y}_{g,K}
=
\frac{1}{K_g}\sum_{j\in U_K(g)}Y_j,
\qquad K_g=\min(K,n_g),
\label{eq:avg}
\end{equation}
where $U_K(g)$ is a uniformly sampled subset of $K_g$ siblings from $I(g)$. With the full group, \avg\ gives the empirical conditional Bellman target for $g$; with a uniform subset, it is an unbiased finite-population estimator of that target.

\model\ instead samples from an empirical full-outcome distribution for each exact state-action group. It stores counts over tuples
\begin{equation}
Z=(s',r,d_{\mathrm{term}},d_{\mathrm{trunc}})
\end{equation}
and samples a full outcome tuple when the corresponding exact group is selected. Sampling the joint tuple preserves dependencies between reward, next state, termination, and truncation. \model\ is directly applicable only in settings where exact state-action groups and full outcome tables are meaningful. For exact groups in the same buffer, \model\ and \samp\ have the same conditional outcome distribution; their different priority-refresh rules can lead to different learning trajectories. Both \avg\ and \model\ use the group-aware IS weights in Equation~\eqref{eq:groupweights}, normalized by their minibatch maximum.

\paragraph{Priority refresh.}
Each method refreshes priorities after the optimizer step using the pre-update TD error $\delta$: $p=|\delta|+\varepsilon$. PER refreshes its anchor; \samp\ refreshes the trained sibling; \model\ refreshes the anchor using its resampled outcome. \avg\ refreshes the anchor or the siblings used in its target average. Appendix~\ref{app:refresh} specifies the complete refresh rules.

\subsection{Approximate latent siblings}
The exact methods use $\kappa(i)=(s_i,a_i)$. For continuous, high-dimensional discrete, or image-based domains, exact state equality is usually too sparse or semantically brittle to define useful sibling groups. We therefore also consider a fixed abstraction $\phi$ and define latent keys
\begin{equation}
\kappa_\phi(i)=(\phi(s_i),a_i).
\end{equation}
In the function-approximation setting considered here, $\phi$ is a frozen VQ-VAE encoder trained on random-policy observations \citep{VanDenOord2017}. It maps each observation to a packed discrete code grid. The same finite-buffer identities apply relative to these fixed abstract groups, but the correction no longer corresponds to the exact conditional outcome distribution for a true state-action pair. We select the abstraction using support diagnostics: code usage, observed $(\phi(s),a)$ buckets, bucket-size quantiles, and the fraction of transitions in sufficiently supported buckets. Appendix~\ref{app:vq} and Table~\ref{tab:vqvae} report the support diagnostics and selected abstractions.

\section{Experiments}
We first evaluate sibling-aware replay in tabular stochastic environments with exact state-action groups, then test approximate sibling groups in MinAtar. We report evaluation return and success, normalized area under the evaluation curve (AUC), and late-window performance. AUC summarizes learning efficiency over training; late-window metrics summarize performance near the end of training. We also measure conditional outcome distortion and descriptive replay statistics, defined in Appendix~\ref{app:metrics}. Exact-group experiments use ten seeds; MinAtar uses five. All methods use mean-squared TD error; Appendix~\ref{app:huber} explains the distinction from Huber loss. Appendix~\ref{app:details} gives complete environment, training, learning-rate, and priority-update details, while Appendix~\ref{app:code} gives code and reproducibility information.

\paragraph{Tabular environments.}
We use two custom diagnostic environments and one gridworld family. OutlierBandit has one state and two actions: a safe action gives $2$, while a risky action gives $100$ with probability $0.01$ and $0$ otherwise. The risky action has lower expected value ($1$ versus $2$), but its rare reward can produce large TD errors. TwoChains has a safe branch returning $1$ after eleven steps and a risky branch returning $80$ with probability $0.01$ after two steps, otherwise $0$. FrozenLake uses the Gym/Gymnasium gridworld interface \citep{Brockman2016,Towers2024Gymnasium}, an $8\times8$ slippery map, intended-action probability $0.99$, step reward $-0.01$, and a 200-step limit. FrozenLake-H100/H300 uses goal reward $+100$ and hole reward $-300$; FrozenLake-H50 uses $+50/-100$ as a milder setting.

Appendix~\ref{app:grid} gives the complete training and evaluation settings.

\paragraph{MinAtar.}
MinAtar provides compact Atari-like visual environments with lower computational cost than full Atari \citep{Young2019}. We use Asterix, Breakout, Freeway, Seaquest, and SpaceInvaders. Clean MinAtar is a control condition. The tail condition is \mpTail: for every positive training reward $r$, we replace it by $r-30$ with probability $0.02$ and by $r+0.02\cdot30/0.98$ otherwise. This preserves the expected reward while introducing rare high-magnitude negative outcomes. Evaluation is always performed on the unperturbed environment. Latent \samp\ uses frozen VQ-VAE state codes paired with actions as approximate sibling keys.

\subsection{Exact state-action groups}
\begin{figure}[t]
  \centering
  \includegraphics[width=0.98\linewidth]{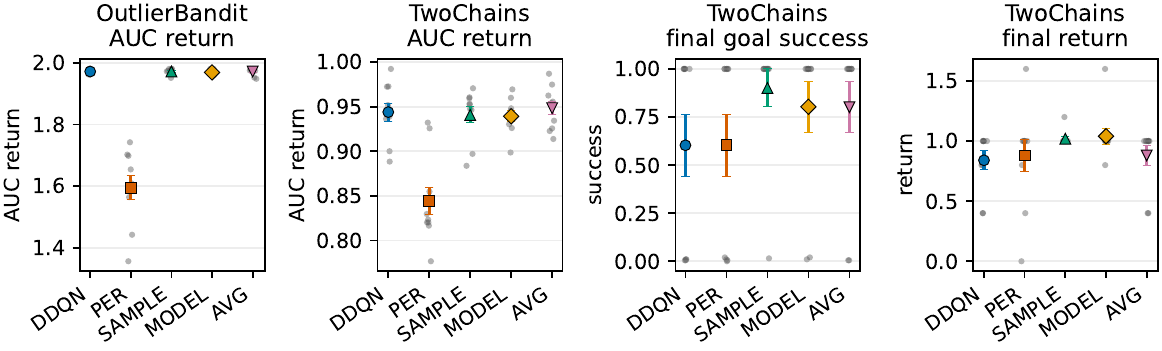}
  \caption{Exact-group results ($\alpha=0.6$, annealed $\beta$). Grey points show seeds; larger method-specific markers show mean $\pm$ SE. Each method uses ten seeds. TwoChains success counts goals from either branch.}
  \label{fig:exacttoy}
\end{figure}

Figure~\ref{fig:exacttoy} and Table~\ref{tab:exactsummary} compare the methods with $\alpha=0.6$ and $\beta$ annealed from $0.4$ toward $1$. In OutlierBandit, all methods eventually learn the safe action, but PER has substantially lower AUC. In TwoChains, \samp, \avg, and \model\ achieve higher mean final goal success than PER. Success includes goals reached through either branch. Section~\ref{sec:ablation} examines sensitivity to the replay parameters.

\begin{table}[t]
\centering
\small
\caption{Tabular-environment metrics, mean $\pm$ standard error over ten seeds. For OutlierBandit and TwoChains, the final columns are final evaluation metrics. For FrozenLake-H100/H300, the final columns are late-window metrics over the final $25$k environment steps.}
\label{tab:exactsummary}
\begin{tabular}{llrrrr}
\toprule
Env. & Method & AUC ret. & AUC succ. & Late/final ret. & Late/final succ. \\
\midrule
OutlierBandit & DDQN & $1.97\se 0.00$ & $0.98\se 0.00$ & $2.00\se 0.00$ & $1.00\se 0.00$ \\
 & PER & $1.59\se 0.04$ & $0.61\se 0.03$ & $2.00\se 0.00$ & $1.00\se 0.00$ \\
 & SAMPLE & $1.97\se 0.00$ & $0.98\se 0.00$ & $2.00\se 0.00$ & $1.00\se 0.00$ \\
 & MODEL & $1.97\se 0.00$ & $0.98\se 0.00$ & $2.00\se 0.00$ & $1.00\se 0.00$ \\
 & AVG & $1.97\se 0.00$ & $0.98\se 0.00$ & $2.00\se 0.00$ & $1.00\se 0.00$ \\
\midrule
TwoChains & DDQN & $0.94\se 0.01$ & $0.66\se 0.05$ & $0.84\se 0.08$ & $0.60\se 0.16$ \\
 & PER & $0.84\se 0.02$ & $0.19\se 0.06$ & $0.88\se 0.13$ & $0.60\se 0.16$ \\
 & SAMPLE & $0.94\se 0.01$ & $0.68\se 0.03$ & $1.02\se 0.02$ & $0.90\se 0.10$ \\
 & MODEL & $0.94\se 0.01$ & $0.66\se 0.03$ & $1.04\se 0.07$ & $0.80\se 0.13$ \\
 & AVG & $0.95\se 0.01$ & $0.69\se 0.03$ & $0.88\se 0.08$ & $0.80\se 0.13$ \\
\midrule
FrozenLake & DDQN & $8.81\se 2.99$ & $0.44\se 0.02$ & $67.71\se 3.83$ & $0.863\se 0.015$ \\
 & PER & $-8.81\se 4.69$ & $0.15\se 0.03$ & $55.72\se 7.58$ & $0.617\se 0.075$ \\
 & SAMPLE & $4.69\se 3.86$ & $0.40\se 0.02$ & $72.10\se 3.71$ & $0.871\se 0.016$ \\
 & MODEL & $12.89\se 3.54$ & $0.43\se 0.02$ & $72.08\se 5.17$ & $0.883\se 0.018$ \\
 & AVG-g2 & $2.64\se 4.13$ & $0.36\se 0.03$ & $69.22\se 4.10$ & $0.832\se 0.021$ \\
\bottomrule
\end{tabular}

\end{table}

\begin{figure}[t]
  \centering
  \includegraphics[width=0.92\linewidth]{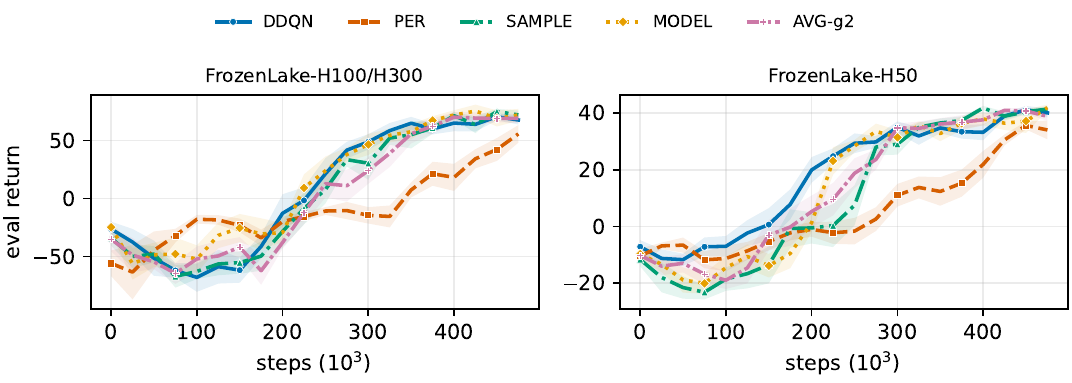}
  \caption{FrozenLake evaluation return curves. H100/H300 is the main exact gridworld test; H50 is the milder reward setting. Shaded regions show standard error over ten seeds.}
  \label{fig:frozencurves}
\end{figure}

In FrozenLake-H100/H300, PER learns more slowly under this schedule (Figure~\ref{fig:frozencurves}). \samp\ and \model\ recover much of its lost late-window return and success (Table~\ref{tab:exactsummary}). Compared with DDQN, \samp\ has higher mean late return and similar late success, while DDQN has higher mean AUC. Thus, the relative performance depends on whether we measure learning efficiency or the final training window.

To measure conditional outcome selection directly, we pool sibling entries that realize the same outcome $o$. Let $q_g(o)=\sum_{i\in I(g):\,Z_i=o}u_i/s_g$ be PER's conditional outcome probability and $f_g(o)=n_{g,o}/n_g$ its empirical buffer frequency. Their total variation (TV) distance is
\begin{equation}
D_g^{\rm out}=\frac12\sum_o|q_g(o)-f_g(o)|.
\label{eq:outcome-tv}
\end{equation}
\samp\ has $D_g^{\rm out}=0$ on any fixed buffer. Figure~\ref{fig:outcome-grid} reports this metric across the parameter sweep; Appendix~\ref{app:metrics} gives the aggregation details and distinguishes outcome distortion, entry concentration, and IS-weight dispersion.

\subsection{Approximate siblings in MinAtar}
\begin{figure}[t]
  \centering
  \includegraphics[width=0.94\linewidth]{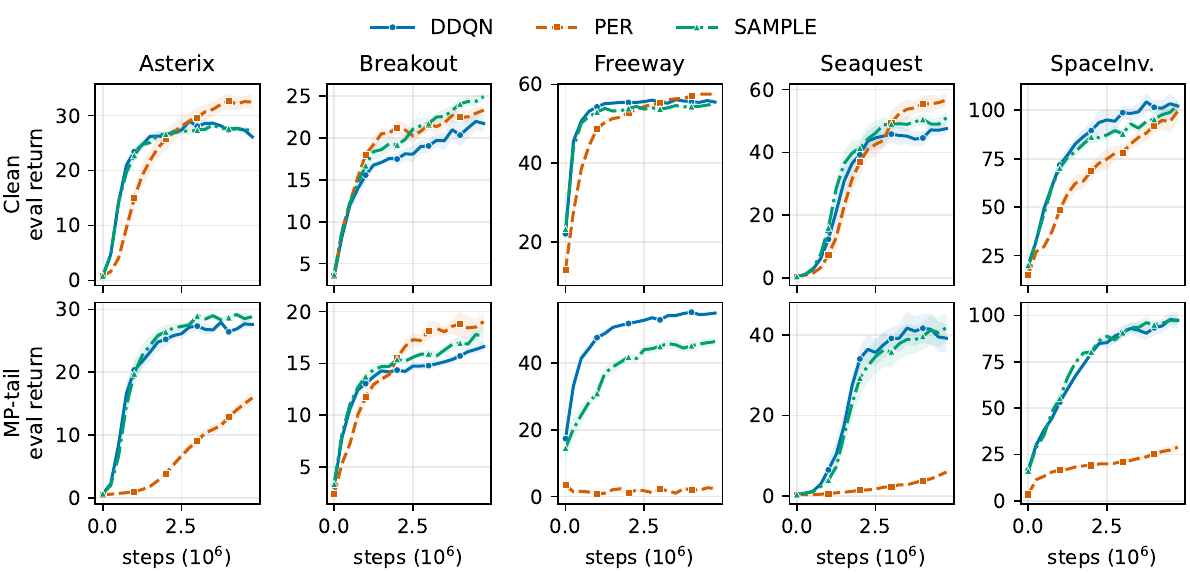}
  \caption{MinAtar evaluation return curves for clean training (top) and \mpTail training (bottom). Evaluation always uses the unperturbed environment; the mean-preserving tail is applied only during training. Shaded regions show standard error over five seeds.}
  \label{fig:minatarcurves}
\end{figure}

To test transfer without tuning specifically for the tail perturbation, we select each method's learning rate on clean games and keep it fixed under \mpTail. Table~\ref{tab:minatarauc} reports the resulting AUC values, and Figure~\ref{fig:minatarcurves} compares clean and \mpTail\ learning curves. Under \mpTail, latent \samp\ improves over PER in Asterix, Freeway, Seaquest, and SpaceInvaders. The effect is largest in Freeway and SpaceInvaders, where PER largely fails under the tail perturbation while \samp\ remains close to the stronger baselines. In Breakout, PER and \samp\ are close on AUC and PER has the higher final return. The benefits therefore vary across games.

\begin{table}[t]
\centering
\small
\caption{MinAtar \mpTail\ AUC return, mean $\pm$ standard error over five seeds. Last columns report paired differences relative to SAMPLE, so positive values indicate that SAMPLE has higher AUC.}
\label{tab:minatarauc}
\begin{tabular}{lrrrrr}
\toprule
Game & DDQN & PER & SAMPLE & S--PER & S--DDQN \\
\midrule
Asterix & $22.07\se 0.75$ & $6.69\se 0.39$ & $22.80\se 0.64$ & $+16.11\se 0.78$ & $+0.73\se 0.50$ \\
Breakout & $13.72\se 0.14$ & $14.69\se 0.23$ & $14.56\se 0.21$ & $-0.12\se 0.14$ & $+0.85\se 0.31$ \\
Freeway & $49.14\se 0.49$ & $1.80\se 0.10$ & $38.57\se 0.90$ & $+36.77\se 0.89$ & $-10.57\se 0.94$ \\
Seaquest & $26.74\se 1.55$ & $2.12\se 0.12$ & $25.10\se 2.39$ & $+22.98\se 2.37$ & $-1.64\se 2.71$ \\
SpaceInv. & $73.87\se 1.72$ & $19.62\se 0.63$ & $75.64\se 0.67$ & $+56.02\se 1.04$ & $+1.77\se 1.51$ \\
\bottomrule
\end{tabular}

\end{table}

Clean MinAtar provides a control condition (Figure~\ref{fig:minatarcurves}, top; Table~\ref{tab:minatar-clean}). PER remains competitive with DDQN in Breakout and Seaquest, while DDQN is stronger in Freeway and SpaceInvaders. \samp's relative performance also varies across clean games. The \mpTail\ condition adds high-magnitude stochastic rewards while preserving their mean, allowing us to examine how the methods respond to reward tails.

DDQN is a strong baseline: under \mpTail, the paired 95\% $t$ intervals do not establish an AUC difference between \samp\ and DDQN in Asterix, Breakout, Seaquest, or SpaceInvaders. In Freeway, the interval favors DDQN. Appendix Table~\ref{tab:minatar-ci} reports the paired differences and intervals for every game and comparator. Together with the exact-group results, these comparisons support \samp{} as a way to retain prioritized group selection while correcting conditional outcome frequencies. Appendix~\ref{app:minatar-diag} reports the descriptive replay diagnostics.

\subsection{Sensitivity to replay parameters}
\label{sec:ablation}
\begin{figure}[t]
  \centering
  \includegraphics[width=0.98\linewidth]{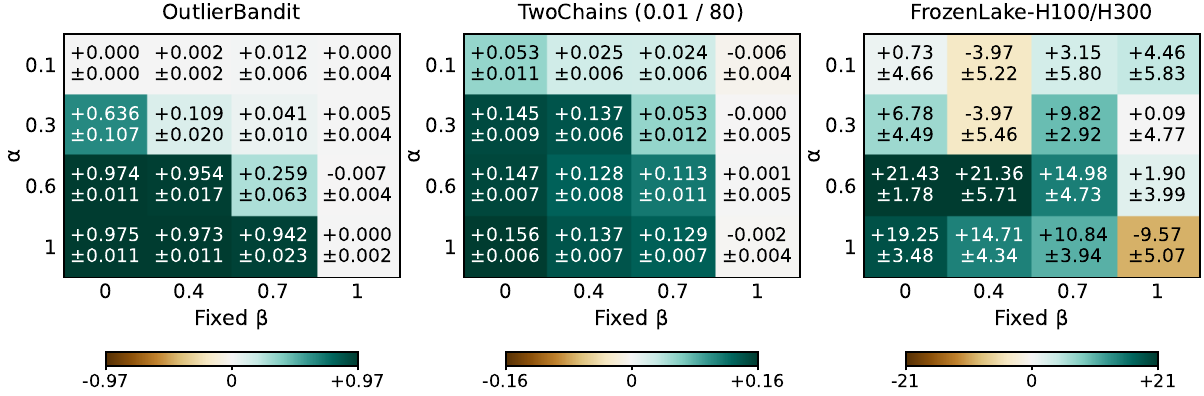}
    \caption{Matched \samp--PER AUC differences, mean $\pm$ SE of ten paired seeds at all 48 fixed $(\alpha,\beta)$ cells across three environments. Positive values favor \samp. Each cell prints the mean and SE; shading scales are normalized separately for each environment.}
  \label{fig:matched-grid}
\end{figure}

We test whether ordinary PER tuning mitigates the performance gap by comparing PER and \samp\ at every combination of $\alpha\in\{0.1,0.3,0.6,1.0\}$ and fixed $\beta\in\{0,0.4,0.7,1.0\}$ in all three exact environments. Each pair uses the same seed and training settings. Figure~\ref{fig:matched-grid} summarizes ten paired seeds per setting; Appendix~\ref{app:grid} gives the full settings and all 48 comparisons.

\samp\ shows clear gains under partial correction in Bandit and TwoChains. At $(\alpha,\beta)=(0.6,0.4)$, the paired \samp--PER AUC differences are $0.954\se0.017$ in Bandit and $0.128\se0.008$ in TwoChains. At $(0.6,1)$ they shrink to $-0.007\se0.004$ and $0.001\se0.005$. Across both environments, $\beta=1$ largely closes the gap; $\alpha=0.1$ also does so in Bandit. Tuning can therefore recover similar mean performance while \samp\ directly controls conditional selection.

FrozenLake is more variable: the paired AUC difference is $21.36\se5.71$ at $(0.6,0.4)$ and $1.90\se3.99$ at $(0.6,1)$, with a reversal of $-9.57\se5.07$ at $(1,1)$. \samp\ has negative mean AUC throughout the $\alpha=1$ row. The simultaneous 95\% interval at $(0.6,0.4)$ is $[-5.75,48.48]$, so it does not establish whether \samp\ or PER has higher mean AUC. Conditional correction does not remove sensitivity to group allocation and update weighting.

\begin{figure}[t]
  \centering
  \includegraphics[width=0.98\linewidth]{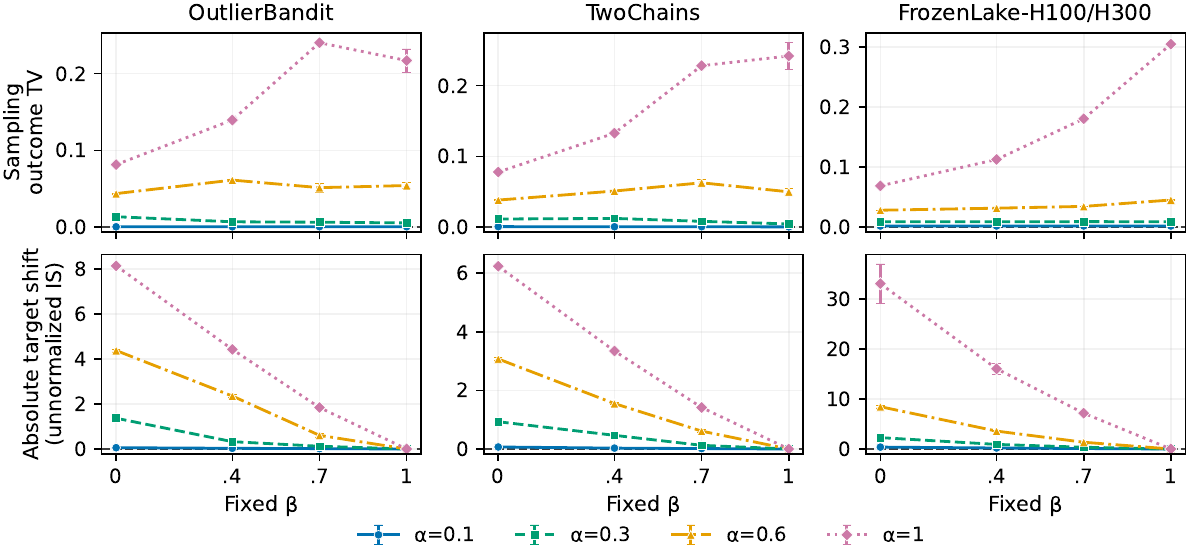}
  \caption{Full-buffer grid diagnostics, mean $\pm$ SE over ten seeds. Styled curves show PER at each $\alpha$; the horizontal dashed line marks SAMPLE's zero-distortion identity. Top: outcome TV. Bottom: absolute conditional target shift under unnormalized IS, distinct from minibatch-max normalization. Both average groups by sampling mass $s_g/S$, then checkpoints within seeds.}
  \label{fig:outcome-grid}
\end{figure}

The outcome measurements in Figure~\ref{fig:outcome-grid} clarify how tuning affects the mechanism. Larger $\alpha$ produces greater mean outcome distortion in these runs. Increasing $\beta$ reduces the target shift under unnormalized IS, reaching zero at $\beta=1$, while sampling distortion persists. This agrees with the fixed-buffer distinction: $\beta$ changes update weights, whereas \samp\ directly controls conditional selection. Appendix~\ref{app:metrics} gives the aggregation details and measurements under the implemented weight normalization.

\section{Discussion and conclusion}
We studied a conditional failure mode of PER: transition-level priorities can over-select stochastic outcomes inside a state-action group. Our finite-buffer decomposition separates group allocation from sibling selection, allowing sibling-aware replay to retain prioritized group selection while recovering empirical outcome frequencies. Exact-group experiments and direct outcome measurements support this mechanism, and approximate MinAtar groups show its practical potential. The parameter sweep qualifies the performance gains: ordinary tuning can close the Bandit and TwoChains gaps, while FrozenLake remains more variable. \samp\ controls conditional selection directly, independently of the chosen $\alpha/\beta$.

The theory characterizes the replay-induced update distribution for a fixed buffer, priorities, and target parameters. It isolates conditional outcome selection as one component of learning dynamics, alongside priority refresh, coverage, exploration, optimization, and abstraction quality.

\samp\ is the most practical variant. It is cheap, preserves outcome variability, and works with exact or latent groups. \model\ is strong in tabular environments because it samples a joint empirical outcome tuple, but it requires a faithful outcome table. \avg\ is theoretically natural and has lower conditional target variance for fixed targets. In FrozenLake, increasing the averaging subset does not consistently improve learning (Appendix~\ref{app:additional}); the fixed-target variance calculation alone does not determine performance.

The value of sibling correction depends on the groups and their stochastic outcomes. If groups are mostly singletons, outcome variability is small, or large errors mainly reflect reducible underfitting, uniform sibling selection may offer little benefit. Approximate groups can also merge observations with different outcome laws or split observations that could share useful experience. The mixed Breakout and clean-MinAtar results motivate studying these conditions. Learning sibling abstractions and combining noise-aware group priorities with conditional outcome correction remain promising directions.

\clearpage
\label{page:references}
\bibliographystyle{plainnat}
\bibliography{references}

\clearpage
\appendix

\section{Code and reproducibility}
\label{app:code}
Code and experiment scripts are available at \url{https://github.com/oscar-omlf/sibling-aware-replay}.

\section{Additional derivations}
\label{app:derivations}

\subsection{Partial importance correction and the conditional exponent}
We analyze one learning step to separate replay sampling from IS weighting. We hold the buffer, sibling groups, and positive priorities fixed, and average over replay sampling at the current network parameters. Gradients pass through predicted Q-values; priorities, IS weights, and Bellman targets are treated as constants. With $u_i=p_i^\alpha$ and unnormalized weights, transition $i$ contributes to the expected gradient with coefficient
\[
P(i)w_i=N^{-\beta}S^{\beta-1}u_i^{1-\beta}.
\]
Normalizing these coefficients within group $g$ gives the effective weights $\pi_\beta(i\mid g)$ in Equation~\eqref{eq:effective-exponent} and mean target $\sum_{i\in I(g)}\pi_\beta(i\mid g)Y_i$. These weights describe the update; PER still samples sibling $i$ with probability $u_i/s_g$.

The total expected-update weight of a group is the sum of its transition coefficients. For PER,
\[
W_{\mathrm{PER}}(g)
=
N^{-\beta}S^{\beta-1}
\sum_{i\in I(g)}u_i^{1-\beta}.
\]
For \samp, each sibling is trained with probability $q_j=s_g/(Sn_g)$. Summing $q_j(Nq_j)^{-\beta}$ over the $n_g$ siblings gives
\[
W_{\mathrm{SAMPLE}}(g)
=
N^{-\beta}S^{\beta-1}
n_g^\beta s_g^{1-\beta}.
\]
The factor $N^{-\beta}S^{\beta-1}$ is common to both methods, but the remaining group terms generally differ when $0<\beta<1$: PER raises each transition mass before summing, whereas \samp\ raises the summed group mass. Thus, PER and \samp\ select groups with the same probabilities at a fixed buffer but can give those groups different total update weights. At $\beta=1$, both unnormalized estimators recover the uniform-buffer expected loss or semi-gradient. This single-step equality does not imply unbiased optimal values or identical Adam updates.

We use stratified sampling: divide the cumulative priority mass into equal intervals and sample once from each. This preserves the expected minibatch mean under PER. We then divide each IS weight by the largest weight in its minibatch. This sampled maximum generally changes the expectation above; at batch size one, for example, every normalized weight equals one regardless of $\beta$. Equation~\eqref{eq:effective-exponent} thus describes unnormalized weights. For \samp, weights and their normalization depend only on the selected groups, so they preserve uniform selection within each group.

\subsection{Effective sampling distribution of SAMPLE}
A transition $j$ can be selected for training whenever the PER anchor lies in its group $g=\kappa(j)$. Multiplying the group probability $s_g/S$ by the uniform sibling probability $1/n_g$ gives Equation~\eqref{eq:samplelaw}; comparing this with the uniform-buffer probability $1/N$ gives the IS ratio in Equation~\eqref{eq:groupweights}.

With full, unnormalized IS correction and fixed priorities and losses, the expected weighted loss equals the uniform-buffer average. Partial correction leaves different weights across groups, but siblings remain uniform within each selected exact group.

\subsection{Bias of PER's within-group target}
For a fixed group $g$, let $Y_i$ denote the Bellman target of sibling $i$. The empirical conditional target is
\[
\bar Y_g = \frac{1}{n_g}\sum_{i\in I(g)}Y_i,
\]
whereas PER without IS correction ($\beta=0$) uses the priority-weighted conditional mean
\[
\tilde Y_g^{\mathrm{PER}}=\sum_{i\in I(g)}\frac{u_i}{s_g}Y_i .
\]
The finite-buffer within-group bias is therefore
\[
B_g^{\mathrm{PER}}=
\tilde Y_g^{\mathrm{PER}}-\bar Y_g
=
\sum_{i\in I(g)}\left(\frac{u_i}{s_g}-\frac{1}{n_g}\right)Y_i .
\]
This term is zero if priorities are constant within the group or if the priority deviations are uncorrelated with sibling targets in the finite buffer. It can be large when rare stochastic outcomes have systematically larger TD errors. This is the sampling-target bias ($\beta=0$); for unnormalized partial correction, replace $u_i/s_g$ by $\pi_\beta(i\mid g)$.

\subsection{Variance of SAMPLE and AVG}
For a selected group $g$, hold its Bellman targets fixed and consider the randomness of sibling selection. \samp\ draws one sibling uniformly, so its target variance is the variance across the stored sibling targets. If \avg\ averages $K$ independent sibling targets sampled with replacement, then
\[
\mathrm{Var}(\widehat Y_{g,K}\mid g)=\frac{\sigma_g^2}{K},
\]
where $\sigma_g^2$ is the variance of a uniformly sampled sibling target. If it samples without replacement, the finite-population correction gives
\[
\mathrm{Var}(\widehat Y_{g,K}\mid g)
=
\frac{\sigma_g^2}{K}\left(1-\frac{K-1}{n_g-1}\right),
\qquad K\le n_g,\quad n_g>1.
\]
A singleton group has zero target variance under both methods. Thus \avg\ has the same conditional mean as \samp\ and no greater fixed-target variance. This calculation alone does not determine learning performance: target networks, stored priorities, and optimizer dynamics also evolve during training.

\subsection{MODEL targets}
\model\ counts occurrences of each complete outcome tuple
\[
Z=(s',r,d_{\mathrm{term}},d_{\mathrm{trunc}})
\]
for each exact $(s,a)$ group in the current buffer. It samples a tuple in proportion to its count, preserving dependencies between rewards, next states, and terminations. Sampling a next state separately from an averaged reward can create an outcome absent from the buffer, rather than sampling from its empirical conditional distribution.

\subsection{Huber loss and mean-value estimation}
\label{app:huber}
Sampling correction and loss choice affect different parts of an update \citep{Fujimoto2020}. Consider fitting a single value $q$ to OutlierBandit's risky rewards: $R=100$ with probability $0.01$ and $R=0$ otherwise. Weight these outcomes by their true probabilities, without priority reweighting. Squared loss has minimizer $\E[R]=1$. Huber loss $H_c$ \citep{Huber1964} has derivative $H_c'(e)=\operatorname{clip}(e,-c,c)$. For threshold $c=1$ and $0\le q\le1$,
\[
\frac{d}{dq}\E[H_1(q-R)]
=0.99q-0.01.
\]
The convex expected loss therefore has unique minimizer $q_{\rm H}=1/99\approx0.0101$.

Huber bounds gradient contributions without changing outcome frequencies; its optimum can differ from the mean for asymmetric targets. The discrepancy depends on the threshold relative to reward scale. Uniform sibling selection recovers empirical outcome frequencies, but choosing Huber can still shift the fitted value away from their mean. Both squared-loss and Huber optima here prefer the safe action's deterministic reward $2$, so this example concerns value estimation rather than policy failure.

\section{Experimental details}
\label{app:details}
\subsection{Priority updates}
\label{app:refresh}
After each learning step, we update stored priorities using TD errors computed before the optimizer step, with the current online and target networks. We assign $p_i=|\delta_i|+10^{-6}$ and sample using masses $p_i^\alpha$. New replay-buffer entries receive the largest raw priority seen so far (initially one); entries otherwise keep their priorities until updated. Group masses $s_g$ track the sum of the current entries' masses, including insertions and evictions.

The \emph{anchor} is the transition initially selected by PER. PER trains on it and updates its priority with its own TD error. \samp\ trains on a uniformly selected sibling, which can be the anchor, and updates only that sibling's priority.

\model\ samples a complete $(s',r,d_{\rm term},d_{\rm trunc})$ outcome from its group's buffer counts. It updates the anchor's priority with the TD error for this sampled outcome, which may differ from the stored one. For PER, \samp, and \model, repeated assignments to an entry follow minibatch order: the last assignment wins.

\avg\ averages all sibling targets or, when capped, a uniform subset sampled without replacement. The anchor is eligible for the selection but is not always included. The priority is the absolute TD error against the averaged target, plus $10^{-6}$. With \texttt{update\_all\_siblings=false}, only the anchor is updated; with the flag true, every sibling used in the average is updated. If several AVG updates address the same entry, the largest proposed priority wins.

OutlierBandit AVG uses full groups with the flag false; TwoChains AVG uses full groups with the flag true; FrozenLake AVG-g2/g4 averages at most two/four siblings with the flag true. These rules allow group priorities to evolve differently across methods.

\paragraph{Loss and optimizer.}
All DQN variants use Double-DQN targets \citep{VanHasselt2016}, Adam \citep{KingmaBa2015}, and mean-squared TD error. MSE retains sensitivity to large residuals, whereas Huber loss \citep{Huber1964} bounds the derivative with respect to the scalar TD error. We hold MSE fixed across methods; the empirical comparison therefore concerns this loss, while the fixed-buffer sampling identities do not depend on it.

\paragraph{MinAtar settings and learning rates.}
For MinAtar \citep{Young2019}, we use two convolutional layers with $16$ and $32$ channels and a fully connected layer with $128$ units. We train for $5$ million environment steps with $\gamma=0.99$, sticky-action probability $0.1$, replay capacity $100{,}000$, batch size $32$, and target-network updates every $1000$ steps. Epsilon decreases linearly from $1$ to $0.1$ over $100{,}000$ steps. PER and \samp\ use $\alpha=0.6$ and linearly anneal $\beta$ from $0.4$ toward $1$ over $5$ million replay updates.

We select a learning rate for each game and method using AUC on clean games, then use it for both clean and \mpTail\ training. Following prior MinAtar work, which sweeps step sizes in powers of two \citep{Young2019}, we test a factor-of-two grid around $3\cdot10^{-5}$: $1.2\cdot10^{-4}$, $6\cdot10^{-5}$, $3\cdot10^{-5}$, $1.5\cdot10^{-5}$, and $7.5\cdot10^{-6}$, plus $3.75\cdot10^{-6}$ and $1.875\cdot10^{-6}$ for PER. Table~\ref{tab:lr-sweep} lists the selected rates. Each reported setting uses five seeds. This comparison measures how well learning rates chosen on clean games transfer to reward tails. For MinAtar AUC differences, unadjusted paired 95\% $t$ intervals use four degrees of freedom and assume approximately normal seed differences. Table~\ref{tab:minatar-ci} reports all paired differences and intervals.

\begin{table}[h]
\centering
\small
\caption{Selected MinAtar learning rates used for the reported clean and \mpTail\ runs.}
\label{tab:lr-sweep}
\begin{tabular}{lrrr}
\toprule
Game & DDQN lr & PER lr & SAMPLE lr \\
\midrule
Asterix & 7.5e-06 & 1.88e-06 & 7.5e-06 \\
Breakout & 6e-05 & 3e-05 & 6e-05 \\
Freeway & 7.5e-06 & 1.88e-06 & 7.5e-06 \\
Seaquest & 7.5e-06 & 3.75e-06 & 7.5e-06 \\
SpaceInv. & 3e-05 & 7.5e-06 & 3e-05 \\
\bottomrule
\end{tabular}

\end{table}

\begin{table}[h]
\centering
\small
\caption{Paired MinAtar \mpTail\ AUC differences over five seeds. Values are mean [unadjusted paired 95\% $t$ interval]. Positive values favor SAMPLE.}
\label{tab:minatar-ci}
\begin{tabular}{lrr}
\toprule
Game & SAMPLE--PER & SAMPLE--DDQN \\
\midrule
Asterix & $+16.11\;[13.93,\,18.28]$ & $+0.73\;[-0.66,\,2.11]$ \\
Breakout & $-0.12\;[-0.51,\,0.26]$ & $+0.85\;[-0.01,\,1.70]$ \\
Freeway & $+36.77\;[34.31,\,39.23]$ & $-10.57\;[-13.19,\,-7.95]$ \\
Seaquest & $+22.98\;[16.40,\,29.55]$ & $-1.64\;[-9.16,\,5.88]$ \\
SpaceInv. & $+56.02\;[53.14,\,58.90]$ & $+1.77\;[-2.42,\,5.97]$ \\
\bottomrule
\end{tabular}
\end{table}

\paragraph{FrozenLake rewards and transitions.}
In FrozenLake-H100/H300, the intended action occurs with probability $0.99$; the two side slips share the remaining probability. Rewards are $+100$ at the goal, $-300$ at holes, and $-0.01$ per step. H50 uses $+50$, $-100$, and $-0.01$. Groups are exact $(s,a)$ pairs. Under the known transition law and infinite-horizon optimal values, with $\gamma=0.99$, $\alpha=0.6$, and no IS correction, the largest downward target shifts among visited greedy state-action pairs with hole risk are approximately $-50.4$ and $-15.3$, respectively. These are analytical examples using the stated reference values.

\subsection{Fixed-\texorpdfstring{$\alpha/\beta$}{alpha/beta} sweep: settings and complete results}
\label{app:grid}
We train PER and \samp\ using Double DQN, two hidden layers of $64$ ReLU units, Adam, MSE, gradient-norm clipping at $10$, batch size $32$, and one gradient step per environment step. Both methods use exact groups, add $10^{-6}$ to absolute TD errors for priorities, and divide IS weights by their minibatch maximum. We stop bootstrapping at time limits. In each cell, the methods use the same training settings and seeds, with fixed $\alpha$ and $\beta$ as listed in Section~\ref{sec:ablation}.

\begin{table}[htbp]
\centering\small
\caption{Training and evaluation settings for the fixed-$\alpha/\beta$ sweep. Evaluation intervals count training episodes; all other intervals count environment steps.}
\label{tab:grid-settings}
\begin{tabular}{@{}lrrr@{}}
\toprule
Setting & OutlierBandit & TwoChains & FrozenLake-H100/H300 \\
\midrule
Training steps & $50{,}000$ & $100{,}000$ & $500{,}000$ \\
Replay capacity & $5{,}000$ & $5{,}000$ & $50{,}000$ \\
Learning starts (steps) & $500$ & $1000$ & $5{,}000$ \\
Learning rate & $7.5\times10^{-4}$ & $7.5\times10^{-4}$ & $3\times10^{-3}$ \\
Discount $\gamma$ & $0$ & $0.99$ & $0.99$ \\
Target update interval & $500$ & $1000$ & $1000$ \\
Initial $\epsilon$ & $1$ & $1$ & $1$ \\
Final $\epsilon$ & $0.02$ & $0.05$ & $0.05$ \\
$\epsilon$ decay steps & $10{,}000$ & $50{,}000$ & $500{,}000$ \\
Evaluation episodes & $500$ & $200$ & $50$ \\
Evaluation interval & $500$ & $100$ & $20$ \\
\bottomrule
\end{tabular}
\end{table}

Table~\ref{tab:grid-settings} gives the environment-specific settings. TwoChains has a safe-chain length of $10$ (eleven steps to reward), risky success probability $0.01$, risky reward $80$, and episode limit $20$. Evaluation is scheduled by episode count, so paired methods can be evaluated at different step counts; each run also has a final evaluation. The main exact-environment comparisons use the same training and evaluation settings, with $\alpha=0.6$ and $\beta$ increasing from $0.4$ toward $1$ over $50$k replay updates in Bandit, $100$k in TwoChains, and $500$k in FrozenLake. H50 uses the FrozenLake settings with its milder rewards. These comparisons use seeds $0$--$9$ in Bandit and TwoChains and $1$--$10$ in FrozenLake.

Normalized return AUC is the main metric. Late return and success average evaluations in the final $25$k steps. We subtract PER from \samp\ at each cell and seed, then compute means and uncertainty across seeds. Tables~\ref{tab:grid-bandit}, \ref{tab:grid-twochains}, and~\ref{tab:grid-frozenlake} report all 48 cells with paired seeds $0$--$9$. The sweep includes 960 runs and 480 paired differences. The outcome diagnostics use these same runs.

To account for the 48 comparisons, we compute 95\% Bonferroni simultaneous $t$ intervals for paired return-AUC differences, with $9$ degrees of freedom and critical value $t_{9,1-0.05/(2\cdot48)}$. These assume approximately normal seed differences. The paired-difference columns report these simultaneous intervals. Figures show descriptive means and SE across seeds; we do not select a best configuration. The sweep holds $\beta$ fixed, whereas Table~\ref{tab:exactsummary} uses annealed $\beta$.
\begin{table}[t]
\centering\small
\caption{OutlierBandit fixed grid, seeds $0$--$9$. PER and SAMPLE: mean $\pm$ SE. Paired SAMPLE minus PER: mean [95\% simultaneous CI across all 48 comparisons].}
\label{tab:grid-bandit}
\begin{tabular}{ccrrr}
\toprule
$\alpha$ & $\beta$ & PER AUC & SAMPLE AUC & Difference [95\% CI] \\
\midrule
0.1 & 0 & $1.978\se 0.001$ & $1.978\se 0.001$ & $0.000\;[0.000,\,0.000]$ \\
0.1 & 0.4 & $1.974\se 0.003$ & $1.976\se 0.002$ & $0.002\;[-0.007,\,0.011]$ \\
0.1 & 0.7 & $1.961\se 0.006$ & $1.974\se 0.002$ & $0.012\;[-0.017,\,0.042]$ \\
0.1 & 1 & $1.973\se 0.003$ & $1.973\se 0.001$ & $0.000\;[-0.018,\,0.018]$ \\
0.3 & 0 & $1.341\se 0.107$ & $1.977\se 0.002$ & $0.636\;[0.129,\,1.143]$ \\
0.3 & 0.4 & $1.866\se 0.020$ & $1.976\se 0.002$ & $0.109\;[0.014,\,0.205]$ \\
0.3 & 0.7 & $1.934\se 0.011$ & $1.974\se 0.003$ & $0.041\;[-0.009,\,0.090]$ \\
0.3 & 1 & $1.960\se 0.004$ & $1.965\se 0.005$ & $0.005\;[-0.016,\,0.026]$ \\
0.6 & 0 & $1.003\se 0.011$ & $1.977\se 0.001$ & $0.974\;[0.921,\,1.027]$ \\
0.6 & 0.4 & $1.019\se 0.017$ & $1.973\se 0.003$ & $0.954\;[0.871,\,1.036]$ \\
0.6 & 0.7 & $1.714\se 0.062$ & $1.974\se 0.003$ & $0.259\;[-0.038,\,0.557]$ \\
0.6 & 1 & $1.977\se 0.001$ & $1.970\se 0.004$ & $-0.007\;[-0.025,\,0.010]$ \\
1 & 0 & $1.003\se 0.011$ & $1.978\se 0.001$ & $0.975\;[0.923,\,1.027]$ \\
1 & 0.4 & $1.003\se 0.011$ & $1.976\se 0.002$ & $0.973\;[0.920,\,1.027]$ \\
1 & 0.7 & $1.033\se 0.023$ & $1.975\se 0.002$ & $0.942\;[0.832,\,1.052]$ \\
1 & 1 & $1.974\se 0.003$ & $1.975\se 0.002$ & $0.000\;[-0.010,\,0.011]$ \\
\bottomrule
\end{tabular}

\end{table}
\begin{table}[t]
\centering\small
\caption{TwoChains fixed grid ($p_{\rm success}=0.01$, $r_{\rm high}=80$), seeds $0$--$9$. PER and SAMPLE: mean $\pm$ SE. Paired SAMPLE minus PER: mean [95\% simultaneous CI across all 48 comparisons].}
\label{tab:grid-twochains}
\begin{tabular}{ccrrr}
\toprule
$\alpha$ & $\beta$ & PER AUC & SAMPLE AUC & Difference [95\% CI] \\
\midrule
0.1 & 0 & $0.904\se 0.012$ & $0.957\se 0.007$ & $0.053\;[0.001,\,0.105]$ \\
0.1 & 0.4 & $0.923\se 0.009$ & $0.948\se 0.007$ & $0.025\;[-0.006,\,0.056]$ \\
0.1 & 0.7 & $0.926\se 0.008$ & $0.949\se 0.004$ & $0.024\;[-0.003,\,0.050]$ \\
0.1 & 1 & $0.949\se 0.006$ & $0.942\se 0.008$ & $-0.006\;[-0.027,\,0.015]$ \\
0.3 & 0 & $0.810\se 0.009$ & $0.955\se 0.006$ & $0.145\;[0.104,\,0.186]$ \\
0.3 & 0.4 & $0.812\se 0.009$ & $0.949\se 0.006$ & $0.137\;[0.108,\,0.166]$ \\
0.3 & 0.7 & $0.888\se 0.015$ & $0.941\se 0.008$ & $0.053\;[-0.005,\,0.111]$ \\
0.3 & 1 & $0.947\se 0.005$ & $0.947\se 0.004$ & $-0.000\;[-0.025,\,0.025]$ \\
0.6 & 0 & $0.806\se 0.008$ & $0.953\se 0.004$ & $0.147\;[0.115,\,0.179]$ \\
0.6 & 0.4 & $0.807\se 0.009$ & $0.935\se 0.006$ & $0.128\;[0.090,\,0.165]$ \\
0.6 & 0.7 & $0.829\se 0.014$ & $0.942\se 0.008$ & $0.113\;[0.061,\,0.164]$ \\
0.6 & 1 & $0.937\se 0.007$ & $0.938\se 0.008$ & $0.001\;[-0.024,\,0.025]$ \\
1 & 0 & $0.804\se 0.008$ & $0.961\se 0.005$ & $0.156\;[0.129,\,0.184]$ \\
1 & 0.4 & $0.807\se 0.009$ & $0.945\se 0.007$ & $0.137\;[0.104,\,0.171]$ \\
1 & 0.7 & $0.811\se 0.008$ & $0.940\se 0.006$ & $0.129\;[0.097,\,0.161]$ \\
1 & 1 & $0.943\se 0.008$ & $0.942\se 0.009$ & $-0.002\;[-0.021,\,0.017]$ \\
\bottomrule
\end{tabular}

\end{table}
\begin{table}[t]
\centering\small
\caption{FrozenLake-H100/H300 fixed grid, seeds $0$--$9$. PER and SAMPLE: mean $\pm$ SE. Paired SAMPLE minus PER: mean [95\% simultaneous CI across all 48 comparisons].}
\label{tab:grid-frozenlake}
\begin{tabular}{ccrrr}
\toprule
$\alpha$ & $\beta$ & PER AUC & SAMPLE AUC & Difference [95\% CI] \\
\midrule
0.1 & 0 & $8.864\se 3.255$ & $9.594\se 4.138$ & $0.730\;[-21.407,\,22.868]$ \\
0.1 & 0.4 & $15.055\se 3.014$ & $11.089\se 3.301$ & $-3.966\;[-28.780,\,20.849]$ \\
0.1 & 0.7 & $11.055\se 3.557$ & $14.208\se 3.824$ & $3.153\;[-24.418,\,30.725]$ \\
0.1 & 1 & $11.694\se 3.086$ & $16.151\se 3.223$ & $4.457\;[-23.250,\,32.163]$ \\
0.3 & 0 & $6.051\se 2.395$ & $12.832\se 4.006$ & $6.780\;[-14.535,\,28.096]$ \\
0.3 & 0.4 & $10.281\se 4.391$ & $6.310\se 4.108$ & $-3.971\;[-29.898,\,21.957]$ \\
0.3 & 0.7 & $6.417\se 1.744$ & $16.236\se 3.095$ & $9.818\;[-4.034,\,23.670]$ \\
0.3 & 1 & $8.370\se 4.086$ & $8.464\se 3.646$ & $0.094\;[-22.551,\,22.739]$ \\
0.6 & 0 & $-13.257\se 2.126$ & $8.175\se 2.419$ & $21.432\;[12.958,\,29.906]$ \\
0.6 & 0.4 & $-10.655\se 2.428$ & $10.709\se 4.272$ & $21.364\;[-5.751,\,48.479]$ \\
0.6 & 0.7 & $-9.196\se 2.468$ & $5.787\se 3.603$ & $14.983\;[-7.481,\,37.448]$ \\
0.6 & 1 & $2.755\se 1.715$ & $4.654\se 3.958$ & $1.899\;[-17.070,\,20.868]$ \\
1 & 0 & $-25.633\se 1.937$ & $-6.380\se 3.910$ & $19.252\;[2.714,\,35.790]$ \\
1 & 0.4 & $-23.983\se 1.273$ & $-9.271\se 3.692$ & $14.712\;[-5.889,\,35.313]$ \\
1 & 0.7 & $-20.707\se 1.505$ & $-9.866\se 3.668$ & $10.841\;[-7.892,\,29.574]$ \\
1 & 1 & $-4.210\se 2.870$ & $-13.782\se 3.536$ & $-9.571\;[-33.657,\,14.514]$ \\
\bottomrule
\end{tabular}

\end{table}

\subsection{Diagnostic definitions}
\label{app:metrics}
\paragraph{Performance Metrics.}
AUC measures learning efficiency over the training horizon $T$. We integrate the evaluation curve by the trapezoidal rule from the first recorded evaluation and divide by $T$; we do not add an unobserved time-zero evaluation. Late-window metrics average evaluation checkpoints near the end of training, with windows specified below.

\paragraph{Variation in importance weights.}
IS-weight ESS is $(\sum_b w_b)^2/\sum_b w_b^2$. For batch size $B$, weight mean $\mu_w$, and population standard deviation $\sigma_w$, it equals $B\mu_w^2/(\mu_w^2+\sigma_w^2)$. Multiplying all weights by a common factor leaves the ESS unchanged. At $\beta=0$, every weight is one and ESS equals $B$, whatever the sampling distribution. ESS therefore measures weight variation, not sampling concentration or target bias.

\paragraph{Outcome frequencies.}
We compare how often an outcome appears among a group's stored transitions, $f_g(o)$, with its probability under replay, $q_g(o)$. Entries with the same complete outcome tuple are counted together. Their total variation distance $D_g^{\rm out}$ (Equation~\eqref{eq:outcome-tv}) is the probability mass that must be reassigned between outcomes to make the distributions agree. This distance ranges from zero to one. With fixed Bellman targets, the target range bounds the resulting mean shift: $|B_g^{\rm PER}|\le(\max_iY_i-\min_iY_i)D_g^{\rm out}$. In Bandit's risky arm, this shift is exactly $100[q_g(100)-f_g(100)]$. Uniform sibling selection has zero outcome TV distance and zero shift relative to its buffer. These quantities describe selection before IS weighting.

\begin{figure}[t]
  \centering
  \includegraphics[width=0.92\linewidth]{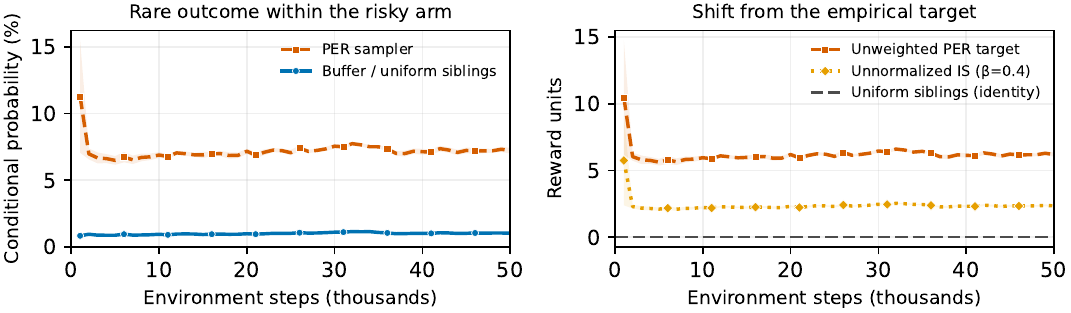}
  \caption{Bandit conditional outcome diagnostics at fixed $\alpha=0.6$, $\beta=0.4$, mean $\pm$ SE over ten seeds. Left: PER's rare-outcome probability versus its buffer frequency. Right: target shifts relative to the empirical mean; the IS curve uses unnormalized weights, not the implemented minibatch-max normalization. Uniform siblings on this same buffer give zero shift by identity.}
  \label{fig:outcome-distortion}
\end{figure}

For Bandit PER at fixed $(\alpha,\beta)=(0.6,0.4)$, we measure the buffer every 1000 training steps using seeds $0$--$9$ of the sweep. Averaging 50 checkpoints per seed gives risky-arm outcome TV distance $0.06197\se0.00090$ and unweighted target shift $6.197\se0.090$ reward units. Unnormalized partial IS reduces this shift to $2.378\se0.065$, using coefficients $u_i^{1-\beta}$. Figure~\ref{fig:outcome-distortion} shows these measurements. Table~\ref{tab:target-shifts} compares the unnormalized calculation with measurements using the implemented weight normalization.

\paragraph{Measurements across the parameter sweep.}
For both methods in all 48 cells, we measure the full buffer after sampling and before optimization or priority updates: every 1000 steps in Bandit and TwoChains and every 2000 steps in FrozenLake, once learning starts. We record counts, priority masses, sampling probabilities, and Double-DQN targets for each joint outcome $(s',r,d_{\rm term},d_{\rm trunc})$. Logging makes no random draws or changes to learning. Performance and diagnostics come from the same ten runs per method and cell.

For each group, we measure how replay shifts the mean target away from the empirical mean $\bar Y_g$. For PER, this shift is $\Delta_g=\sum_o q_g(o)Y_g(o)-\bar Y_g$; after unnormalized IS weighting, $\Delta_{g,\beta}$ uses Equation~\eqref{eq:effective-exponent}. SAMPLE retains $f_g(o)$ before and after group-aware weighting, so both conditional shifts are zero.

We average across groups using their selection probabilities $s_g/S$. We call $\sum_g(s_g/S)D_g^{\rm out}$ the mass-weighted outcome TV distance and $\sum_g(s_g/S)|\Delta_g|$ the mass-weighted absolute target shift. Absolute values prevent shifts in opposite directions from cancelling. We use the same group weights for $\Delta_{g,\beta}$ to isolate the within-group effect of IS. Figure~\ref{fig:outcome-grid} reports these metrics. We first average checkpoints within each seed, then report mean and SE across seeds.

\paragraph{Targets after weight normalization.}
For each training minibatch, we record outcome counts, normalized weight sums, and each target's deviation from its group's empirical mean, $Y_b-\bar Y_{g_b}$. Using the actual normalized weights $\widehat w$, the average across logged steps $t$ and batch entries $b$ is
\[
\frac{\sum_{t,b}\widehat w_{t,b}(Y_{t,b}-\bar Y_{g_b,t})}{\sum_{t,b}\widehat w_{t,b}}.
\]
This signed average includes sampling noise and allows shifts in different groups to cancel. It describes target deviations rather than the full parameter update. We log the full-buffer and training-minibatch target evaluations separately. For \samp, weights are constant within each selected group, so normalization preserves conditional uniformity.

Table~\ref{tab:target-shifts} compares the absolute conditional shifts with the signed measurements under the implemented weights. The latter allow cancellation across groups, so their magnitudes can differ; neither metric alone determines learning performance.

\begin{table}[htbp]
\centering\small
\caption{Target shifts at $\alpha=0.6$, $\beta=0.4$, mean $\pm$ SE over ten seeds. Conditional columns compare group-weighted absolute shifts before and after unnormalized IS; sampled columns use signed target deviations and actual minibatch-normalized weights. SAMPLE's conditional shifts are zero.}
\label{tab:target-shifts}
\begin{tabular}{@{}lrrrr@{}}
\toprule
& \multicolumn{2}{c}{PER conditional shift (absolute)} & \multicolumn{2}{c}{Sampled shift (signed)} \\
Environment & Before IS & After IS & PER & SAMPLE \\
\midrule
OutlierBandit & $6.14\se0.09$ & $2.36\se0.06$ & $2.43\se0.11$ & $-0.0003\se0.0200$ \\
TwoChains & $4.09\se0.06$ & $1.55\se0.03$ & $1.30\se0.06$ & $-0.0007\se0.0079$ \\
FrozenLake & $9.61\se0.19$ & $3.58\se0.07$ & $-0.67\se0.08$ & $-0.048\se0.056$ \\
\bottomrule
\end{tabular}
\end{table}

\paragraph{Sampling individual buffer entries.}
This diagnostic measures unequal sampling of individual buffer entries within a group, even when those entries share the same outcome. For PER's conditional probabilities $q_g(i)=u_i/s_g$, define the within-group entry-concentration statistic
\[
C_g=\sum_{i\in I(g)}\frac{(q_g(i)-1/n_g)^2}{1/n_g}
=n_g\sum_{i\in I(g)}q_g(i)^2-1,
\qquad C=\sum_g\frac{s_g}{S}C_g.
\]
Uniform selection gives $C_g=0$; concentrating probability on fewer entries increases it. Equivalently, $C_g=n_g/n_{{\rm eff},g}-1$, where $n_{{\rm eff},g}=1/\sum_i q_g(i)^2$ is the effective number of sampled entries. This measures concentration within groups, separately from group allocation.

The ratio $R_i=n_g u_i/s_g$ compares an entry's conditional PER probability with uniform sibling sampling. We log its minibatch mean and population standard deviation before priority updates. Under PER, $\E[R_i\mid g]=1+C_g$, so the logged mean minus one estimates $C$, also with stratified sampling. Finite-sample estimates can be negative; we leave them unclipped. We average checkpoints within runs, then compute means and SE across seeds.

Under \samp, the trained sibling is uniform, so the actual conditional divergence is zero. The logged ratio still uses PER probabilities: it describes how PER would sample from \samp's current buffer. Its mean has expectation one, while $\sigma_R^2+(\mu_R-1)^2$ estimates that hypothetical PER divergence, using the logged mean $\mu_R$ and standard deviation $\sigma_R$. For AVG and MODEL, the logged ratio describes anchor priorities; their targets are averaged or resampled.

Unequal entry probabilities do not necessarily distort outcome frequencies. At step $41$k of Bandit PER seed $0$ with annealed $\beta$, the buffer contains $4953$ safe-reward transitions and $47$ zero-reward risky transitions, with no rare reward. Each group has only one outcome, so $D_g^{\rm out}=0$, yet the logged entry-divergence estimate is $4207.8$ and all sampled TD errors are zero. Insertion and priority updates can leave identical outcomes with different priorities. Outcome TV ignores this entry-level variation. For fixed targets, Cauchy--Schwarz gives $(B_g^{\rm PER})^2\le C_g\,\mathrm{Var}_{U_g}(Y)$; a positive upper bound does not establish bias.

\begin{figure}[t]
  \centering
  \includegraphics[width=0.90\linewidth]{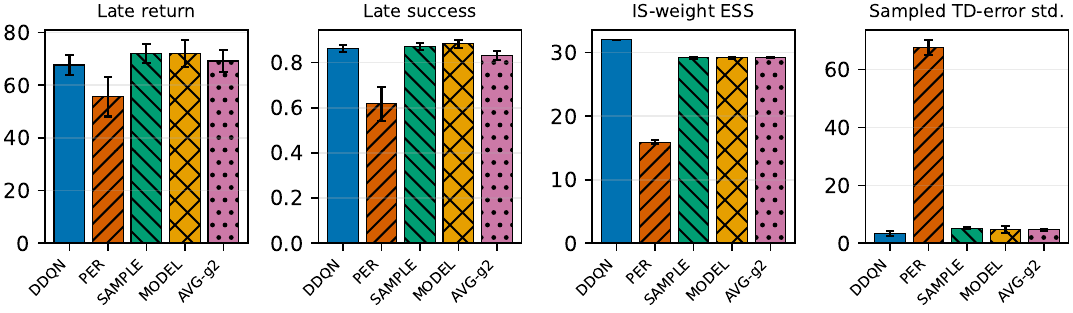}
  \caption{FrozenLake-H100/H300, mean $\pm$ SE over ten seeds. Return and success average the final $25$k steps; IS-weight ESS and sampled TD-error dispersion average the final $100$k. Return and success summarize late-window learning performance.}
  \label{fig:frozenlate}
\end{figure}

\begin{figure}[t]
  \centering
  \includegraphics[width=0.86\linewidth]{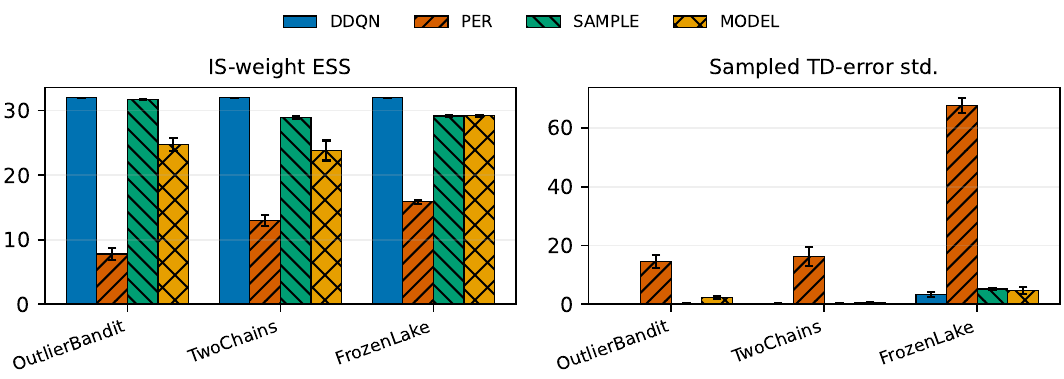}
  \caption{Exact-group descriptive diagnostics, mean $\pm$ SE over ten seeds. IS-weight ESS measures weight dispersion; sampled TD-error standard deviation depends on the sampled or constructed targets. Windows follow Appendix~\ref{app:metrics}.}
  \label{fig:exactdiag}
\end{figure}

We average these statistics over the final $\max(25\mathrm{k},0.2T)$ environment steps: $25$k for OutlierBandit and TwoChains, $100$k for FrozenLake, and $1$M for MinAtar. TD-error standard deviation describes signed residuals in sampled minibatches, not error against the true value function; averaging targets can reduce it directly. Figure~\ref{fig:frozenlate} combines late-window performance and diagnostics for FrozenLake-H100/H300, while Figure~\ref{fig:exactdiag} compares the diagnostic quantities across the exact-group environments. Action L1, reported in Figure~\ref{fig:minatardiag}, compares window averages of \emph{cumulative} behavior and replay action frequencies, not instantaneous sampling probabilities.


\subsection{Other replay diagnostics}
\label{app:minatar-diag}
\begin{figure}[ht!]
  \centering
  \includegraphics[width=0.90\linewidth]{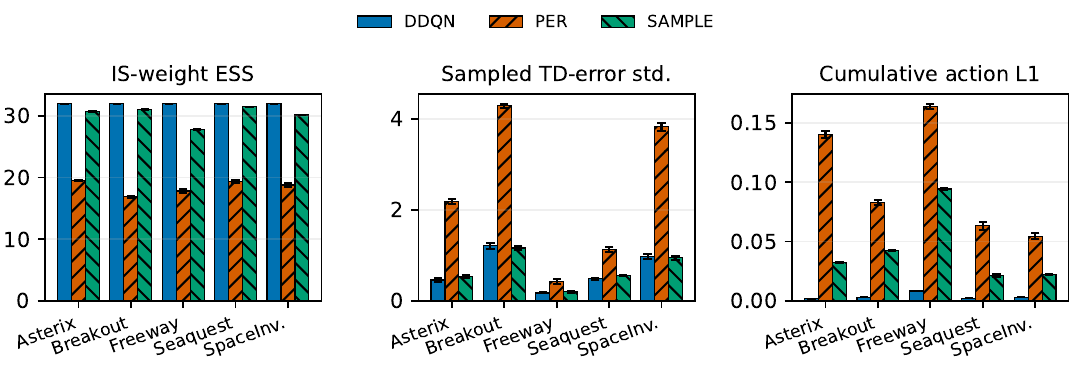}
  \caption{MinAtar \mpTail\ descriptive diagnostics, mean $\pm$ SE over five seeds, final $20\%$ of training. ESS measures IS-weight dispersion; action L1 compares late-window averages of cumulative behavior and replay frequencies.}
  \label{fig:minatardiag}
\end{figure}

Figure~\ref{fig:minatardiag} shows IS-weight ESS, sampled TD-error standard deviation, and the L1 difference between cumulative behavior and replay action frequencies. \samp\ generally has less variable weights and smaller sampled TD errors under \mpTail. The cumulative action counts reflect both past behavior and buffer turnover; they do not measure the current group-selection probabilities in Equation~\eqref{eq:samplelaw}.

\section{VQ-VAE grouping}
\label{app:vq}
We train a separate VQ-VAE for each MinAtar game and freeze its encoder before RL training. For each game, we collect $1.5$ million observations using uniformly random actions, then train for $100{,}000$ minibatch updates with batch size $256$. The encoder maps an observation to a discrete code grid. Observations with the same code grid and action share an approximate sibling group, $(\phi(s),a)$. The VQ-VAE defines these groups and is not trained jointly with the RL objective. We tune the reconstruction loss (MSE or binary cross-entropy, BCE), code-grid size, and codebook size using reconstruction and grouping diagnostics. The selected encoders use BCE.

We choose the abstraction size to provide enough siblings per group while retaining distinctions between states. We compare code grids of $1\times1$ and $2\times1$ and codebooks with $8$, $16$, $32$, and $256$ entries where relevant. Coarse codes can merge states with different outcome distributions; fine codes can leave too few siblings for useful replacement. The selected $2\times1,K=16$ codes provide useful group sizes in four games. SpaceInvaders uses $K=32$ to retain more state distinctions while keeping enough siblings. Table~\ref{tab:vqvae} reports group sizes and code-action coverage.

\begin{table}[ht]
\centering
\small
\caption{Selected VQ-VAE grouping statistics. Recon.: final logged training-minibatch BCE; perp.: exponentiated code-use entropy; pair cov.: fraction of possible code-action buckets observed. Median bucket counts transitions per occupied bucket; frac $\ge100$ is the fraction of sampled transitions in buckets containing at least $100$ transitions.}
\label{tab:vqvae}
\begin{tabular}{llrrrrr}
\toprule
Game & Grid,$K$ & recon. & perp. & pair cov. & median bucket & frac $\ge100$ \\
\midrule
Asterix & $2\times1,16$ & 0.061 & 14.0 & 0.664 & 38 & 0.822 \\
Breakout & $2\times1,16$ & 0.028 & 8.7 & 0.133 & 156 & 0.973 \\
Freeway & $2\times1,16$ & 0.090 & 15.1 & 0.951 & 33 & 0.677 \\
Seaquest & $2\times1,16$ & 0.024 & 12.1 & 0.569 & 44 & 0.767 \\
SpaceInv. & $2\times1,32$ & 0.082 & 16.8 & 0.158 & 41 & 0.760 \\
\bottomrule
\end{tabular}

\end{table}

\section{Control and reward sensitivity results}
\label{app:additional}
\paragraph{Clean MinAtar.}
Table~\ref{tab:minatar-clean} summarizes the clean-training curves in Figure~\ref{fig:minatarcurves}. PER has higher mean AUC than DDQN in Breakout and Seaquest, whereas DDQN leads in Asterix, Freeway, and SpaceInvaders. \samp's relative performance also varies by game. These controls distinguish the response to stochastic reward tails from differences already present under clean training.

\begin{table}[ht]
    \centering
    \small
    \caption{Clean MinAtar AUC return, mean $\pm$ standard error over five seeds.}
    \label{tab:minatar-clean}
    \begin{tabular}{lrrr}
\toprule
Game & DDQN & PER & SAMPLE \\
\midrule
Asterix & $23.70\se 0.46$ & $22.97\se 1.34$ & $23.39\se 0.50$ \\
Breakout & $17.23\se 0.62$ & $19.17\se 0.35$ & $19.24\se 0.51$ \\
Freeway & $52.97\se 0.15$ & $49.80\se 0.34$ & $51.70\se 0.11$ \\
Seaquest & $32.63\se 2.11$ & $34.02\se 2.27$ & $35.57\se 1.84$ \\
SpaceInv. & $83.39\se 3.02$ & $67.92\se 3.30$ & $79.18\se 1.45$ \\
\bottomrule
\end{tabular}

\end{table}

\begin{figure}[ht]
  \centering
  \includegraphics[width=0.78\linewidth]{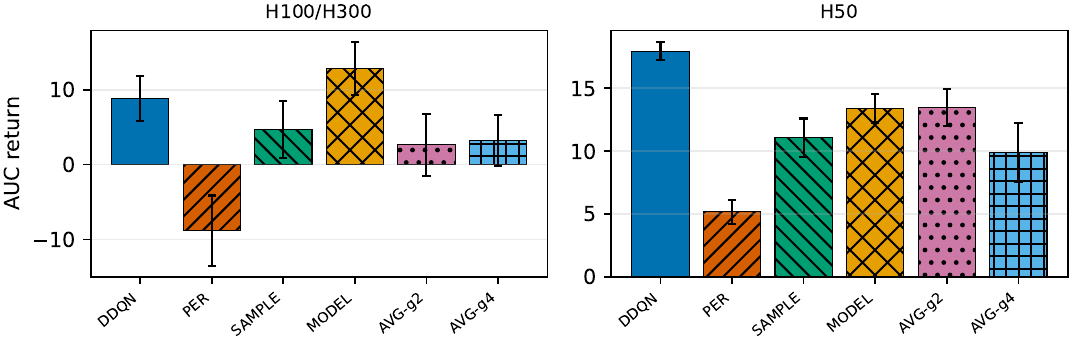}
  \caption{FrozenLake AUC return for both selected reward schedules, mean $\pm$ SE over ten seeds, including AVG-g2 and AVG-g4. H100/H300 is the sharper exact-state stress test; H50 is the milder reward setting.}
  \label{fig:frozenauc}
\end{figure}

\paragraph{Milder FrozenLake rewards.}
Figure~\ref{fig:frozenauc} compares the two reward schedules. In H50, sibling-aware methods have higher mean AUC than PER, but DDQN has the highest mean AUC. Thus, recovery relative to PER also occurs with milder rewards, without implying an advantage over uniform replay. AVG-g4 has slightly higher mean AUC than AVG-g2 in H100/H300 ($3.21$ versus $2.64$), but lower mean AUC in H50 ($9.90$ versus $13.47$); increasing the averaging subset therefore does not consistently improve learning.

\end{document}